\documentclass{article}

\usepackage[preprint]{neurips_2024}

\usepackage[utf8]{inputenc}
\usepackage[T1]{fontenc}
\usepackage{hyperref}
\usepackage{url}
\usepackage{booktabs}
\usepackage{amsfonts}
\usepackage{nicefrac}
\usepackage{microtype}
\usepackage{xcolor}
\usepackage{subcaption}

\usepackage[disable]{todonotes}

\hypersetup{
			colorlinks=true,
			linkcolor=red,
            linktoc=page,
			anchorcolor=black,
			citecolor=gray,
			urlcolor=blue,
			pdftitle={Transformers and slot encoding for sample efficient physical world modeling},
			pdfauthor={Francesco Petri}
}

\newcommand{\R}{\mathbb{R}}
\newcommand{\acronym}{FPTT}
\newcommand{\fullname}{Future-Predicting Transformer Triplet for world modeling}

\def\subsectionautorefname{section}
\def\figureautorefname{figure}
\def\tableautorefname{table}
\newcommand{\Autoref}[1]{%
  \begingroup%
  \def\chapterautorefname{Chapter}%
  \def\sectionautorefname{Section}%
  \def\subsectionautorefname{Section}%
  \def\figureautorefname{Figure}%
  \def\tableautorefname{Table}%
  \autoref{#1}%
  \endgroup%
}

\title{Transformers and slot encoding for sample efficient physical world modelling}

\author{%
  Francesco Petri\thanks{Corresponding author.} \\
  Institute for Cognitive Sciences and Technologies\\
  National Research Council\\
  Rome, Italy \\
  \texttt{francesco.petri@uniroma1.it} \\
  \And
  Luigi Asprino \\
  University of Bologna and ISTC-CNR \\
  Bologna, Italy \\
  \texttt{luigi.asprino@unibo.it} \\
  \And
  Aldo Gangemi \\
  University of Bologna and ISTC-CNR \\
  Bologna, Italy \\
  \texttt{aldo.gangemi@unibo.it} \\
}

\begin{document}

\maketitle

\begin{abstract}
  World modelling, i.e. building a representation of the rules that govern the world so as to predict its evolution, is an essential ability for any agent interacting with the physical world. 
  Recent applications of the Transformer architecture to the problem of world modelling from video input show notable improvements in sample efficiency. However, existing approaches tend to work only at the image level thus disregarding that the environment is composed of objects interacting with each other. 
  In this paper, we propose an architecture combining Transformers for world modelling with the slot-attention paradigm, an approach for learning representations of objects appearing in a scene. 
  We describe the resulting neural architecture and report experimental results showing an improvement over the existing solutions in terms of sample efficiency and a reduction of the variation of the performance over the training examples.
  The code for our architecture and experiments is available at \url{https://github.com/torchipeppo/transformers-and-slot-encoding-for-wm}
\end{abstract}

\section{Introduction}
\label{sec:intro}

World modelling is the ability of an artificial agent to build an internal representation of the world in which it operates.
This representation is employed by the agent to forecast  the evolution of the world.
The problem of building a world model spans many branches of artificial intelligence, such as planning and reinforcement learning \citep{micheli23, paster2021iclr}, physics modelling and reasoning \citep{ding2021nips}, and robotics \citep{wm-in-robotics:wu2022corl}.
An accurate representation of the world allows building simulations that, in turn, enable practitioners to gather additional data and test the performance of an artificial agent without interacting with their environment.
This is convenient because interacting with an agent's environment can be time-consuming, risky due to possible failures of physical components, and sometimes even impossible due to the potential unavailability of the environment (e.g. experimenting with the exploration of Mars by a rover).

Recent applications of the Transformer architecture \citep{transformer:vaswani17} to the task of world modelling from video input suggest that this family of architectures is not only it is capable of capturing the dynamics of the environment,
but it is also capable of learning with high sample efficiency \citep{micheli23, robine2023iclr}.
However, existing approaches typically
operate directly at the image level, with little regard for the objects contained within it.  
Understanding how objects interact with each other and with the environment is of paramount importance, as it endows agents with an intuitive theory of object motion~\citep{mccloskey1983intuitive}.
In neural architecture, this problem is addressed in a separate line of research \citep{slot-attn:locatello2020nips, savi:kipf22, slotformer:wu23}, that focuses on learning object-based representations that allow the objects in the scene and their interactions to be modelled explicitly.
We hypothesise that Transformers may benefit from object-based representations to learn more accurate models of the world.

\paragraph{Problem statement}
In this paper, we focus on physical world modelling through the analysis and prediction of synthetic videos. Learning a model of basic physical laws such as gravity and collision is very important for any agent working in a real environment to understand better the world's evolution and the consequences of the agent's actions.

Machine learning research on modelling intuitive physics aims to replicate the innate understanding of physical concepts that humans display since their first months of age \citep{intphys-1:mccloskey1983, intphys-2:baillargeon2004}. Specifically, we approach the physical interaction output prediction problem, as defined in a recent survey \citep{intphys-survey:jiafei2022ijcai}. In this context, the agent is shown a video, composed of a sequence of frames \(x_1, ..., x_T\) depicting several objects interacting in a world governed by physical laws such as gravity and collision. The agent is then asked to predict the final outcome of the situation, which requires estimating how the objects will behave after what is shown in the input video \citep{clevrer:yi2020iclr}.

\paragraph{Contribution}
We design an entirely transformer-based architecture for world modelling, inspired by the principles of representation learning with slot encoding \citep{savi:kipf22, steve:singh22}. The evaluation is based on how well the learned model predicts the outcome of the situations it gets shown. We show that this allows us to reap the benefits from both approaches (i.e. slot encoding and transformers), while also noticeably improving the stability of the training process.

\paragraph{Paper outline} We structure the rest of the paper as follows. \Autoref{sec:related} discusses some previous works on the topics of world modelling and representation learning. \Autoref{sec:method} describes our architecture and how it was trained. \Autoref{sec:experiments} describes the evaluation experiments we performed. \Autoref{sec:conclusion} offers some final remarks on this work.

\section{Related work}
\label{sec:related}

\subsection{World modelling}

The world modelling problem has received tremendous attention in the last few years. In reinforcement learning, being able to simulate the environment dynamics is especially useful, because it enables the agent to act and learn in its own simulated world without paying the cost of interacting with the "real" environment. The Dreamer algorithm, in its various versions \citep{dreamerv1:hafner2020iclr, dreamerv2:hafner2021iclr, dreamerv3:hafner2023preprint}, has been relatively influential on the topic, with \citet{wm-in-robotics:wu2022corl} being an application in a robotic domain, where real-world interaction has the unfortunate potential of breaking usually costly equipment, in addition to time costs.
Additional approaches in reinforcement learning include solutions based on causal discovery and reasoning \citep{yu2023ijcai}, which aim at learning a causal model of the environment to better understand the interactions between the agent and the world, and even provide an explanation for the actions taken by the agent.
Transformer-based approaches are also studied, due to their generally good performance in different tasks and the sample efficiency they provide in this specific problem \citep{micheli23, robine2023iclr}.

In the case of agents acting in a real physical environment, learning a model of the basic physical laws of the world is essential to act effectively in the environment and understand the consequences of each move. 
We refer to this problem as intuitive physics modelling.
For this reason, several solutions have been studied for this problem with approaches ranging from deep learning \citep{physics-rpin:qi2021iclr}, to violation of expectation \citep{physics-voe:Piloto2022IntuitivePL}, to causal reasoning \citep{physics-causal:li2022aaai}.

With this work, we aim to improve the general level of performance and stability of Transformer-based approaches by implementing an unsupervised representation learning module, specifically one based on the principles of slot encoding.

\subsection{Object-oriented representation learning}

In this work, we integrate concepts from the line of research on learning object-oriented representations from images and videos \citep{slot-attn:locatello2020nips, parts:zoran2021iccv, jia2023iclr}. In particular, slot encoders for video \citep{savi:kipf22, steve:singh22} learn a representation that tracks the prominent objects in a video frame by frame. The structure of our architecture is based on that of slot encoders, but we try to streamline it by focusing exclusively on using Transformer modules, which allows us to convert the image to just one intermediate form, i.e. a sequence of tokens to be elaborated by the Transformers, while \citet{steve:singh22} requires two separate elaborations of the image: one to produce convolutional features and one to produce a sequence of tokens.

The idea of leveraging slot encoding mechanisms for world modelling has also been explored in \citet{slotformer:wu23}. However, while that work uses a single transformer as the dynamics modelling module, we experiment with the idea of keeping representation correction and dynamics advancement separate, where each step is learned by a different, smaller neural model.

This is not the only existing approach: an additional, earlier line of work focuses on using generative models with object-centric features for images \citep{air:eslami2016nips, genesis:engelcke2020iclr} and video \citep{sqair:kosiorek2018nips} to distinguish the objects present in a scene and improve image/video generation with the learned object awareness. \citet{scalor:jiang2020iclr} applies this paradigm to world modelling. Other approaches include spatial attention \citep{space:lin2020iclr} and latent space factorization \citep{simone:kabra2021nips}.

\section{Method}
\label{sec:method}

\begin{figure}
    \centering
    \includegraphics[width=\textwidth]{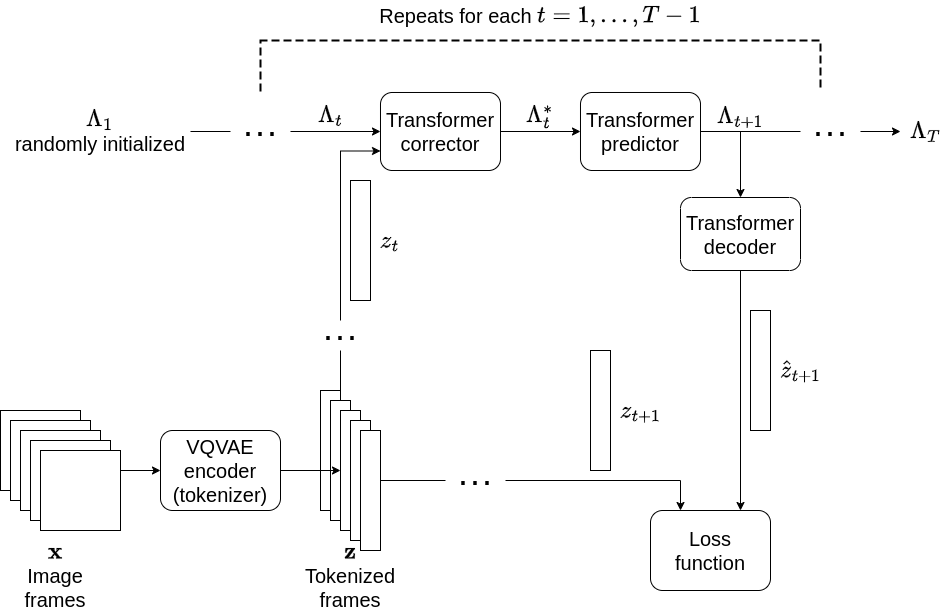}
    \caption{Architecture diagram for the \fullname{}.}
    \label{fig:arch-diagram}
\end{figure}

We introduce a multi-stage architecture, called "\fullname{}" (\textbf{\acronym{}}), which aims to model the behaviour of objects in a set of videos so as to predict their evolution.

We frame the world modelling problem as a sequence learning one.
We use transformers as the fundamental building block of our architecture, due to their proven performance in world modelling~\citep{micheli23} as well as other tasks that can be reduced to modelling and manipulating a sequence of tokens \citep{transformer:vaswani17, taming:esser21}.%
In particular, we leverage the recent work by \citet{micheli23}, which showcases a sample-efficient application of transformers to world modelling.

We also borrow some ideas from the slot-attention mechanism \citep{savi:kipf22, steve:singh22} which learns a compact representation for objects appearing in a video.

\subsection{Notation}
We indicate with \(x_t\) the \(t\)-th frame of a video and with \(z_t\) a sequence of tokens corresponding to \(x_t\). \(\Lambda_t(x)\) is the internal representation
of the input video \(\mathbf{x}\) up to time \(t\), i.e. given frames \(x_1, ..., x_{t-1}\), while \(\Lambda_{t}^{*}(x)\) is the "corrected" representation which also includes information from frame \(x_t\).
The initial representation \(\Lambda_1(x)\)
is randomly determined for initialization purposes.

\subsection{Architectural overview}

\Autoref{fig:arch-diagram} shows the high-level components of
\acronym{}.
Further details on the implementation, e.g. the hyperparameters of the architecture, can be found in Appendix \ref{sec:hypars}.

The architecture takes as input a sequence of \(T\) frames of a video, i.e.  \(x_t\) with \(t = 1, ..., T\).
The frames are processed sequentially, so that \(\Lambda_{t+1}(x)\) is determined by combining the previous representation \(\Lambda_t(x)\) with the new frame \(x_t\).

As in previous works~\citep{micheli23}, each frame \(x_t\) is transformed into a corresponding sequence of tokens \(z_t\) by a discrete Vector Quantized Variational Autoencoder (VQVAE)
\citep{vqvae:oord17, taming:esser21}, as transformers need to work on sequences of tokens.
Further details on the VQVAE can be found in \autoref{sss:tokenizer}.

After this preliminary step, the sequence of tokens \(z_t\) is processed by the core components of the architecture which are meant to predict the next representation \(\Lambda_{t+1}(x)\) based on the current one \(\Lambda_{t}(x)\) and \(z_t\). These components, both based on the transformer architecture,
have the same high-level purpose as their counterparts in slot attention for video \citep{savi:kipf22, steve:singh22} architectures:

\begin{itemize}
    \item The \textbf{corrector transformer} (see \autoref{sec:corrector}) which compares the previous (internal) representation \(\Lambda_t(x)\) with the tokenized representation of the current frame \(z_t\)
    in order to consistently align the internal representation with the actual evolution of the video;
    
    \item The \textbf{predictor transformer} (see \autoref{sec:predictor}) which predicts the evolution of the world state and produces the representation of the next time step \(\Lambda_{t+1}(x)\) on the basis of the result of the corrector, i.e., \(\Lambda_{t}^{*}(x)\). \(\Lambda_{t+1}(x)\) is then passed to the corrector for the next stage.    
\end{itemize}

The result of the prediction at stage \(t\), i.e. \(\Lambda_{t+1}(x)\), is also passed to the \textbf{decoder transformer} (see \autoref{sec:decoder}).
This component transforms the predicted internal representation  \(\Lambda_{t+1}(x)\) into a sequence of tokens \(\hat z_{t+1}\).
Finally, the loss is calculated by comparing \(\hat{z}_{t+1}\) with \(z_{t+1}\), i.e. the sequence of tokens obtained from the input frame.
All the above steps (correction, prediction, decoding and loss calculation) are computed for each input frame except the last one, i.e. for \(t = 1, ..., T-1\). The last frame is not processed at training time because it would require the existence of a frame \(x_{T+1}\) to calculate the loss against, which is impossible to provide since the video only goes up to frame \(x_T\) by definition.

\subsection{Vector Quantized Variational Autoencoder for tokenization}
\label{sss:tokenizer}

The Vector Quantized Variational Autoencoder (VQVAE) transforms video frames into a format that subsequent transformers can process. This format is a sequence of \(L\) tokens, with each token represented by a vector of the space \(\mathcal{V} = \left\{ v_1, v_2, ..., v_N \right\} \subset \R^d\), where \(d\) is defined as a hyperparameter of the architecture (see appendix \ref{sec:hypars}).

The VQVAE alternates residual convolutional layers, attention blocks, and convolutional downsampling layers to convert an image\footnotemark{} \(x \in \R^{W \times H \times 3}\) (W and H are the width and height of the image, respectively)
to a latent-space representation \(z_l(x) = \left( z_{l,1}(x), z_{l,2}(x), ..., z_{l,L}(x) \right) \in \R^{L \times d}\). Then each latent vector is quantized into a token simply by picking the closest embedding vector in \(\mathcal{V}\), that is to say, \(z(x) \in \R^{L \times d}\) is such that \(z_i(x) = argmin_{v \in \mathcal{V}} \left( || z_{l,i}(x) - v ||_2 \right)\) for each \(i = 1, ..., L\).

\footnotetext{We omit the \(t\) subscript in this paragraph for ease of notation, since the VQVAE processes images as single entities, not as parts of a video.}

A decoder network with a symmetrical structure to the encoder (not shown in \autoref{fig:arch-diagram}) is used to reconvert a token sequence \(z\) back into an image \(\hat x (z)\) for the purposes of training the whole autoencoder pair.

\subsection{Corrector transformer}
\label{sec:corrector}

The purpose of the corrector transformer is to avoid drifting, i.e. making the internal representation stick with the evolution of the video. This is achieved by updating the estimated representation \(\Lambda_t(x)\) with the corresponding frame \textit{\(z_t\)} thus producing a corrected representation \(\Lambda^*_t(x)\).
It is implemented by a transformer that produces the corrected representation (\(\Lambda^*_t(x)\)) by performing an unmasked cross-attention of the two inputs (\(\Lambda_t(x)\) and \(z_t\)).

It is worth noting that this structure fits neither the transformer encoder nor the transformer decoder descriptions as traditionally defined in \citet{transformer:vaswani17}, since we perform cross-attention (like a decoder) without including a causal mask (like an encoder).
This allows us to compare the two input sequences as a whole, without arbitrarily limiting the context.
We also note that the purpose of this transformer is not to perform autoregressive generation, so a non-causal flow of information causes no harm.

\subsection{Predictor transformer}
\label{sec:predictor}
The predictor transformer performs self-attention on the representation \(\Lambda^*_t(x)\) to estimate its advancement to the next time step \(\Lambda_{t+1}(x)\).

The predictor and corrector transformers can be seen as two halves of one model, dedicated to predicting the next internal representation on the basis of the current representation and the current frame of the video being processed. For this reason, each of them has individually fewer layers compared to the decoder (see \autoref{sec:decoder}).

\subsection{Decoder transformer}
\label{sec:decoder}

The decoder transformer
converts a representation \(\Lambda_{t+1}(x)\) into a sequence of tokens \(\hat{z}_{t+1}\). 
The loss is computed by comparing \(\hat{z}_{t+1}\) with \(z_{t+1}\), i.e. the sequence of tokens obtained from the input frame.

As it can be seen in \autoref{fig:arch-diagram}, we position this stage between the predict step that generates \(\Lambda_{t+1}(x)\) and the subsequent correct step. 
Thus, the loss computes the error on the prediction of the frame \(z_{t+1}\).
It is worth noticing that this solution diverges from the original slot-attention architectures~\citep{savi:kipf22, steve:singh22}, which calculate the loss on the output of the corrector.
The calculation of the loss on the predicted representation has the effect of directing the model's attention towards the accurate prediction of future events, rather than towards the representation of the current, thus emphasizing the world modelling objective.

\subsection{Training}

The VQVAE is trained in isolation with respect to the whole architecture.
To enhance the stability of the training process, we maintain a fixed configuration of the VQVAE parameters throughout the training of the other components.
Following \citet{micheli23}, the loss is a combination of a mean absolute error and a perceptual loss \citep{perceptual-loss:johnson2016eccv} on the reconstruction, as well as a commitment loss on the embeddings \citep{vqvae:oord17}.

As for the corrector, predictor and decoder transformers, they are trained together in an end-to-end fashion, with the objective to minimize a cross-entropy loss on the (tokenized versions of the) predicted frames \(\hat z_2, ..., \hat z_T\) with respect to the ground-truth ones \(z_2, ..., z_T\).

Both parts of the architecture are trained in a self-supervised way, with unlabeled videos from a suitable dataset. See \autoref{sub:experimental-setup} for further details on the dataset.

\section{Experiments}
\label{sec:experiments}

The ability to model the world of the proposed architecture (\acronym{}) is assessed through a physical reasoning task (see Section~\ref{sub:auxiliary-problem}) that requires the ability to predict how a set of objects moves in a given environment.
Specifically, we experiment with the PHYRE dataset which provides a benchmark containing a set of simple classical mechanics puzzles in a 2D physical environment~\citep{phyre:bakhtin19}.
The performance of \acronym{} is compared against two baselines.

\subsection{Physical reasoning task}
\label{sub:auxiliary-problem}

We adhere to the definition of a physical reasoning task as outlined in the PHYRE benchmark \citep{phyre:bakhtin19}.
The task is set in a two-dimensional world that simulates simple deterministic Newtonian physics with a constant downward gravitational force and a small amount of friction. This world contains non-deformable objects, distinguished by colour, that can be static (i.e. they remain in a fixed position) or dynamic (i.e. they move if they collide with another object and are influenced by the force of gravity).

\begin{figure}[t]
    \centering
    \begin{subfigure}[b]{0.25\textwidth}
        \centering
        \includegraphics[width=\textwidth]{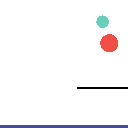}
        \caption{Initial state}
    \end{subfigure}
    \hfill
    \begin{subfigure}[b]{0.25\textwidth}
        \centering
        \includegraphics[width=\textwidth]{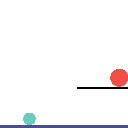}
        \caption{Success}
    \end{subfigure}
    \hfill
    \begin{subfigure}[b]{0.25\textwidth}
        \centering
        \includegraphics[width=\textwidth]{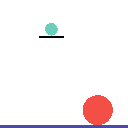}
        \caption{Failure case}
    \end{subfigure}
    \caption{Example frames from the PHYRE dataset.}
    \label{fig:phyre}
\end{figure}

A task consists of an initial world state and a goal (see \autoref{fig:phyre}). 
The initial world state is a pre-defined configuration of objects. 
The goal for all tasks is the following: at the end of the simulation the green object must touch the blue object. 
If the goal is achieved, the task succeeds (as in the PHYRE terminology).

Given video (as a sequence of frames), the objective of the world model is to build an internal representation that can be used to predict if the depicted task will succeed or fail.
The ability to predict
that
represents an auxiliary classification problem. 
This allows us to indirectly assess the performance of the world models.
It is worth noticing that the same evaluation protocol is used in related work~\citep{slotformer:wu23}.

\subsection{Baselines}

The performance of the proposed architecture is evaluated in comparison to the following baselines:

\begin{itemize}
   
    \item \textbf{STEVE} \citep{steve:singh22}, a slot encoding architecture that is also based on the correction-prediction pattern.
    The input frame is encoded using a convolutional neural network (CNN) which feeds a recurrent neural network (RNN) acting as corrector.
    The result is then passed to the predictor, a single-layer transformer, which is then translated into a token sequence (i.e. \(\hat z_t\)) by a transformer decoder.
    \item A single decoder transformer, intended to replicate the approach to world modelling from visual data by \citet{micheli23}. 
    It is worth noticing that \citet{micheli23} work in a reinforcement learning setting, where the state of the world can also be influenced by the actions of an agent.
    Therefore, we adapted this architecture to experiment with a pre-rendered video dataset where no agents interact with the environment.
    Specifically, in \citet{micheli23} the transformer takes as extra inputs and outputs,  the action taken by the agent and the reward given by the environment, which we removed from the architecture to fit it in the physical reasoning problem.
    As a result, in our experiments, a decoder transformer predicts the (tokenized version of the) next frame \(z_{t+1}\) directly from the previous one \(z_t\), without using an internal representation. 
    In the following, we refer to this architecture as \textbf{decoder-only}. 

\end{itemize}

\subsection{Experimental setup}
\label{sub:experimental-setup}

\begin{figure}[t]
    \centering
    \includegraphics[width=\textwidth]{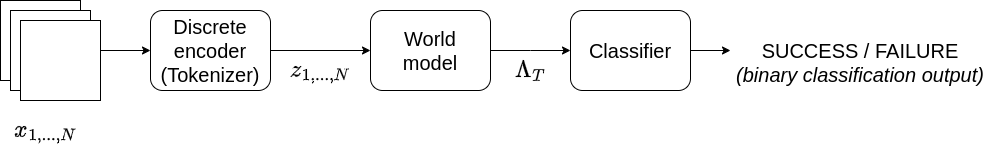}
    \caption{Diagram of the experimental setup, showing how the classifier is positioned with respect to the world modelling architecture. Note: in the case of the decoder-only baseline, replace \(\Lambda_T\) with \(z_T\).}
    \label{fig:experimental_setup}
\end{figure}

We experiment on a dataset of synthetic videos presented in \citet{physics-rpin:qi2021iclr}. This dataset was generated by rendering simulations from the PHYRE benchmark for physical reasoning \citep{phyre:bakhtin19}.
Specifically, we focus exclusively on videos from B-tier tasks, and within-template evaluation. 
\Autoref{fig:experimental_setup} shows the experimental setup. 
The world model takes a video (as a sequence of frames) from the PHYRE task and builds a representation. 
This representation is then passed to a classifier which predicts the result of the task (i.e. success or failure).
As for the classifier, we use a BERT-like encoder architecture \citep{bert:devlin19}, trained in a supervised manner.

\begin{figure}[t]
    \centering
    \includegraphics[width=\textwidth]{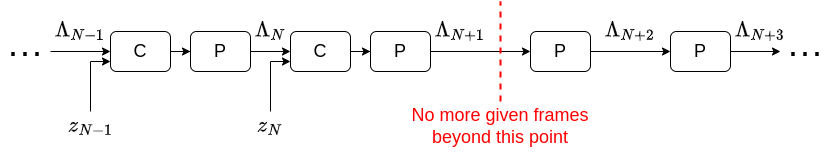}
    \caption{Illustration of the process described in \autoref{sub:experimental-setup} for FPTT and STEVE. Notation has been simplified with respect to \autoref{fig:arch-diagram}. C represents the corrector transformer, P stands for the predictor one.}
    \label{fig:corrpred-vs-monopred}
\end{figure}

As for \acronym{} and STEVE, we proceed as follows. 
Each video in the dataset represents a task that is labeled as either "success" or "failure".
The world model is given the first \(N\) frames of a video whose total length is \(T\) frames, with \(N < T\). The remaining \(T-N\) frames are kept hidden from the model.
In order to obtain the representation of the whole video, including both the given section and an estimation for the following hidden one, the \(N\) given frames are processed as usual, updating the representation in the correct step and advancing it to the next timestep in the predict step.
Afterwards, the remaining \(T-N\) steps are projected by simply repeating the predict step, skipping the correction for the hidden frames (see \autoref{fig:corrpred-vs-monopred}). In the experiments, \(N\) is set to 5, while \(T\) varies depending on each video, ranging from 7 to 18, with many videos being 12-15 frames long.

We follow a similar approach for the decoder-only architecture, accounting for the lack of an internal representation in this case. 
The world model sees the first N frames and the remaining are generated autoregressively by the transformer. 
The final frame, i.e. the sequence of tokens \(z_T\), is passed to the classifier instead of a representation.

All the experimenting architectures employ the same pre-trained VQVAE for transforming each input frame into a sequence of tokens (see \autoref{sss:tokenizer}).

Overall, the dataset contains 50K videos: 47.5K of them (95\%) are used for training purposes, and 2.5K for evaluation (5\%).
We consider the following classification metrics: accuracy, precision, recall, and F1 score.
The experiments were run on a server with an NVIDIA A100 graphics card, with 40GB memory.

Each experiment on \acronym{} and STEVE took about an hour to complete, while experiments on the decoder-only baseline took about double the time. We attribute this to the fact that, without an internal representation, the transformer needs to deal with the longer \(z_t\) sequences, and the time and memory requirements for the self-attention operation scale quadratically with sequence length.

\subsection{Results}

\begin{figure}[t]
\begin{subfigure}[t]{0.49\textwidth}
    \includegraphics[width=\textwidth]{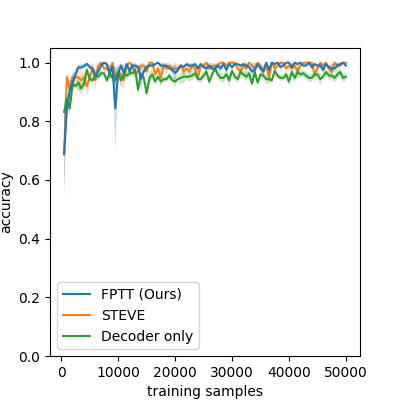}
    \caption{Accuracy}
    \label{fig:accuracy}
\end{subfigure}
\hfill
\begin{subfigure}[t]{0.49\textwidth}
    \includegraphics[width=\textwidth]{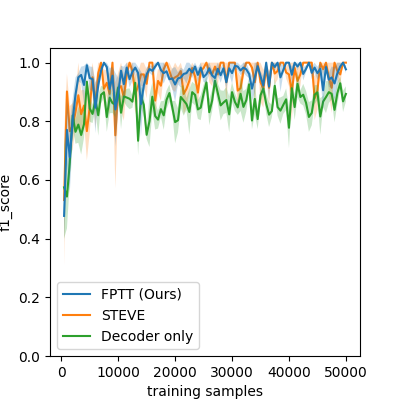}
    \caption{F1 score}
    \label{fig:f1score}
\end{subfigure}
\\
\begin{subfigure}[t]{0.49\textwidth}
    \includegraphics[width=\textwidth]{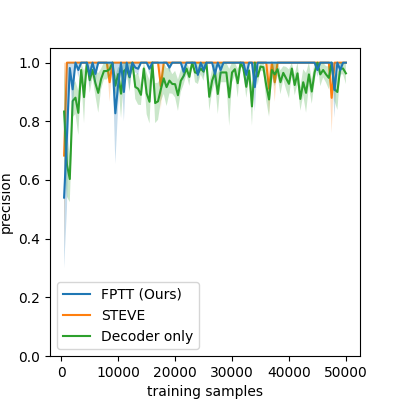}
    \caption{Precision}
    \label{fig:precision}
\end{subfigure}
\hfill
\begin{subfigure}[t]{0.49\textwidth}
    \includegraphics[width=\textwidth]{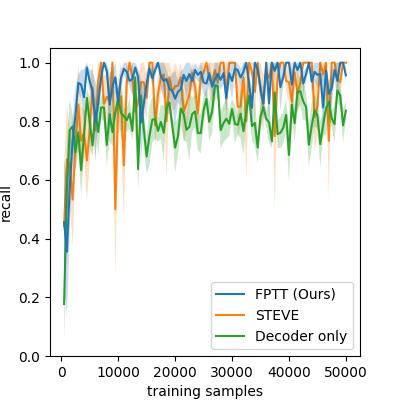}
    \caption{Recall}
    \label{fig:recall}
\end{subfigure}
\caption{Classification results on test data as a function of the number of training samples observed. Each line represents an average over 5 experiments; the coloured bands indicate the standard error of the mean.}
\label{fig:results}
\end{figure}

We report on the result of our experiments in \autoref{fig:results}.
Each experiment was repeated 5 times for statistical significance.

It can be observed that \acronym{} and STEVE are comparable in terms of performance, with the decoder-only exhibiting a lower performance.
Upon closer analysis, especially of the F1 score (\autoref{fig:f1score}) and recall (\autoref{fig:recall}), we observe that  \acronym{} tends to be more stable than STEVE as it shows a narrower error margin. 
We also point out that \acronym{} achieves peak performance earlier than STEVE. 
These considerations allow us to conclude that \acronym{} is a more stable and sample-efficient world model than the baselines.

The only exception to the above is the precision metric (\autoref{fig:precision}), where STEVE achieves 1 after seeing very few examples. 
However, its performance on recall is significantly more erratic and lower on average, resulting in an overall lower F1 score. 

In addition to qualitative analysis, we support the sample efficiency claim with quantitative evidence.
Based on \citet{sampleff-metric:gu2017iclr}, we set a performance threshold at 0.95 on the F1 score and measure the number of training steps required in each experimental run to reach this threshold for the first time.
As a consistency condition, we require the threshold to be exceeded for 4 consecutive training epochs (each epoch has 500 training steps, followed by an evaluation phase.)
We report the result in \autoref{tab:sample-efficiency}, noting that our world model shows a \(35\%\) improvement with respect to the STEVE baseline with about half the error margin.

\begin{table}[b]
\centering
\begin{tabular}{lc}
    \hline
     & Mean \(\pm\) Standard error \\
    \hline
    FPTT (Ours) & \(\mathbf{5500 \pm 758.3}\) \\
    STEVE & \(8500 \pm 1483.2\) \\
    Decoder only & \(29000\) *(see caption) \\
    \hline
\end{tabular}
\caption{Quantitative measure of sample efficiency by reporting the number of training steps required for the F1 score to be above 0.95 consistently for 4 epochs in a row. We report the mean result for each architecture, and the standard error of the mean as a measure of confidence. *Note: For the decoder-only baseline, only one out of the 5 runs cleared the threshold.}
\label{tab:sample-efficiency}
\end{table}

\section{Discussion}
\label{sec:conclusion}

In this section, we discuss the limitations of our approach, outline possible future research directions and draw our conclusions from this study.

\subsection{Limitations}
\label{sub:limitations}

Despite the observed performance improvements,  the representation remains opaque and lacks interpretability. 
Our preliminary attempts at replicating the object segmentation displayed by slot-attention architectures \citep{savi:kipf22, steve:singh22} have not yet yielded positive results. 
However, this may be overturned by more systematic experimentation in the future.

We must also note that, despite the effort to keep the number of layers of the components low, the current version of the \acronym{} architecture is memory-demanding, with a requirement of about 22 GB of GPU RAM at training time, including the tokenizer and the transformer triplet and excluding the classifier.

Finally, we acknowledge that we only experimented on a simplistic synthetic dataset for the initial tests of this new model. 
Although we claim that the presented experiments demonstrate the benefit of the proposed architecture, we do plan to extend the experiments to more complex video datasets such as MOVi-E \citep{movi-e:GreffBBDDDFGGHK22} and Physion \citep{physion:bear2021nips}, which will test its performance in more realistic scenes as well as its generalization capabilities.

\subsection{Conclusion}

We propose a new architecture, the \fullname{} (\acronym{}), which leverages the power of transformers for sequence learning to model the behaviour of objects in a set of videos and predict the evolution of the environment.

We experimentally show that our architecture outperforms transformer-based world models \citep{micheli23}, and improves on slot-attention methods \citep{savi:kipf22, steve:singh22} in terms of sample efficiency and stability during the training process.

In the future, we intend to conduct further experiments with the architecture in more interactive environments, in which objects can be moved by agents. 
Moreover, we would like to study applications to causal discovery problems \citep{yu2023ijcai}, where learning a compact representation that can be interpreted causally might help in understanding complex scenarios.

\begin{ack}
This work has been funded by the Italian National PhD Program on Artificial Intelligence run by Sapienza University of Rome in collaboration with the Italian National Council for Research, and by the Italian PNRR MUR project PE0000013-FAIR.
\end{ack}


{
\small

\bibliographystyle{plainnat}
\bibliography{main.bbl}

\begin{thebibliography}{42}
\providecommand{\natexlab}[1]{#1}
\providecommand{\url}[1]{\texttt{#1}}
\expandafter\ifx\csname urlstyle\endcsname\relax
  \providecommand{\doi}[1]{doi: #1}\else
  \providecommand{\doi}{doi: \begingroup \urlstyle{rm}\Url}\fi

\bibitem[Baillargeon(2004)]{intphys-2:baillargeon2004}
Renée Baillargeon.
\newblock Infants' physical world.
\newblock \emph{Current Directions in Psychological Science}, 13:\penalty0
  89--94, 06 2004.
\newblock \doi{10.1111/j.0963-7214.2004.00281.x}.

\bibitem[Bakhtin et~al.(2019)Bakhtin, van~der Maaten, Johnson, Gustafson, and
  Girshick]{phyre:bakhtin19}
Anton Bakhtin, Laurens van~der Maaten, Justin Johnson, Laura Gustafson, and
  Ross~B. Girshick.
\newblock {PHYRE:} {A} new benchmark for physical reasoning.
\newblock In Hanna~M. Wallach, Hugo Larochelle, Alina Beygelzimer, Florence
  d'Alch{\'{e}}{-}Buc, Emily~B. Fox, and Roman Garnett, editors, \emph{Advances
  in Neural Information Processing Systems 32: Annual Conference on Neural
  Information Processing Systems 2019, NeurIPS 2019, December 8-14, 2019,
  Vancouver, BC, Canada}, pages 5083--5094, 2019.
\newblock URL
  \url{https://proceedings.neurips.cc/paper/2019/hash/4191ef5f6c1576762869ac49281130c9-Abstract.html}.

\bibitem[Bear et~al.(2021)Bear, Wang, Mrowca, Binder, Tung, Pramod, Holdaway,
  Tao, Smith, Sun, Li, Kanwisher, Tenenbaum, Yamins, and
  Fan]{physion:bear2021nips}
Daniel Bear, Elias Wang, Damian Mrowca, Felix~J. Binder, Hsiao{-}Yu Tung, R.~T.
  Pramod, Cameron Holdaway, Sirui Tao, Kevin~A. Smith, Fan{-}Yun Sun, Fei{-}Fei
  Li, Nancy Kanwisher, Josh Tenenbaum, Dan Yamins, and Judith~E. Fan.
\newblock Physion: Evaluating physical prediction from vision in humans and
  machines.
\newblock In Joaquin Vanschoren and Sai{-}Kit Yeung, editors, \emph{Proceedings
  of the Neural Information Processing Systems Track on Datasets and Benchmarks
  1, NeurIPS Datasets and Benchmarks 2021, December 2021, virtual}, 2021.
\newblock URL
  \url{https://datasets-benchmarks-proceedings.neurips.cc/paper/2021/hash/d09bf41544a3365a46c9077ebb5e35c3-Abstract-round1.html}.

\bibitem[Devlin et~al.(2019)Devlin, Chang, Lee, and Toutanova]{bert:devlin19}
Jacob Devlin, Ming{-}Wei Chang, Kenton Lee, and Kristina Toutanova.
\newblock {BERT:} pre-training of deep bidirectional transformers for language
  understanding.
\newblock In Jill Burstein, Christy Doran, and Thamar Solorio, editors,
  \emph{Proceedings of the 2019 Conference of the North American Chapter of the
  Association for Computational Linguistics: Human Language Technologies,
  {NAACL-HLT} 2019, Minneapolis, MN, USA, June 2-7, 2019, Volume 1 (Long and
  Short Papers)}, pages 4171--4186. Association for Computational Linguistics,
  2019.
\newblock \doi{10.18653/V1/N19-1423}.
\newblock URL \url{https://doi.org/10.18653/v1/n19-1423}.

\bibitem[Ding et~al.(2021)Ding, Chen, Du, Luo, Tenenbaum, and
  Gan]{ding2021nips}
Mingyu Ding, Zhenfang Chen, Tao Du, Ping Luo, Josh Tenenbaum, and Chuang Gan.
\newblock Dynamic visual reasoning by learning differentiable physics models
  from video and language.
\newblock In Marc'Aurelio Ranzato, Alina Beygelzimer, Yann~N. Dauphin, Percy
  Liang, and Jennifer~Wortman Vaughan, editors, \emph{Advances in Neural
  Information Processing Systems 34: Annual Conference on Neural Information
  Processing Systems 2021, NeurIPS 2021, December 6-14, 2021, virtual}, pages
  887--899, 2021.
\newblock URL
  \url{https://proceedings.neurips.cc/paper/2021/hash/07845cd9aefa6cde3f8926d25138a3a2-Abstract.html}.

\bibitem[Duan et~al.(2022)Duan, Dasgupta, Fischer, and
  Tan]{intphys-survey:jiafei2022ijcai}
Jiafei Duan, Arijit Dasgupta, Jason Fischer, and Cheston Tan.
\newblock A survey on machine learning approaches for modelling intuitive
  physics.
\newblock In Lud~De Raedt, editor, \emph{Proceedings of the Thirty-First
  International Joint Conference on Artificial Intelligence, {IJCAI-22}}, pages
  5444--5452. International Joint Conferences on Artificial Intelligence
  Organization, 7 2022.
\newblock \doi{10.24963/ijcai.2022/763}.
\newblock URL \url{https://doi.org/10.24963/ijcai.2022/763}.
\newblock Survey Track.

\bibitem[Engelcke et~al.(2020)Engelcke, Kosiorek, Jones, and
  Posner]{genesis:engelcke2020iclr}
Martin Engelcke, Adam~R. Kosiorek, Oiwi~Parker Jones, and Ingmar Posner.
\newblock {GENESIS:} generative scene inference and sampling with
  object-centric latent representations.
\newblock In \emph{8th International Conference on Learning Representations,
  {ICLR} 2020, Addis Ababa, Ethiopia, April 26-30, 2020}. OpenReview.net, 2020.
\newblock URL \url{https://openreview.net/forum?id=BkxfaTVFwH}.

\bibitem[Eslami et~al.(2016)Eslami, Heess, Weber, Tassa, Szepesvari,
  Kavukcuoglu, and Hinton]{air:eslami2016nips}
S.~M.~Ali Eslami, Nicolas Heess, Theophane Weber, Yuval Tassa, David
  Szepesvari, Koray Kavukcuoglu, and Geoffrey~E. Hinton.
\newblock Attend, infer, repeat: Fast scene understanding with generative
  models.
\newblock In Daniel~D. Lee, Masashi Sugiyama, Ulrike von Luxburg, Isabelle
  Guyon, and Roman Garnett, editors, \emph{Advances in Neural Information
  Processing Systems 29: Annual Conference on Neural Information Processing
  Systems 2016, December 5-10, 2016, Barcelona, Spain}, pages 3225--3233, 2016.
\newblock URL
  \url{https://proceedings.neurips.cc/paper/2016/hash/52947e0ade57a09e4a1386d08f17b656-Abstract.html}.

\bibitem[Esser et~al.(2021)Esser, Rombach, and Ommer]{taming:esser21}
Patrick Esser, Robin Rombach, and Bj{\"{o}}rn Ommer.
\newblock Taming transformers for high-resolution image synthesis.
\newblock In \emph{{IEEE} Conference on Computer Vision and Pattern
  Recognition, {CVPR} 2021, virtual, June 19-25, 2021}, pages 12873--12883.
  Computer Vision Foundation / {IEEE}, 2021.
\newblock \doi{10.1109/CVPR46437.2021.01268}.
\newblock URL
  \url{https://openaccess.thecvf.com/content/CVPR2021/html/Esser\_Taming\_Transformers\_for\_High-Resolution\_Image\_Synthesis\_CVPR\_2021\_paper.html}.

\bibitem[Greff et~al.(2022)Greff, Belletti, Beyer, Doersch, Du, Duckworth,
  Fleet, Gnanapragasam, Golemo, Herrmann, Kipf, Kundu, Lagun, Laradji, Liu,
  Meyer, Miao, Nowrouzezahrai, {\"{O}}ztireli, Pot, Radwan, Rebain, Sabour,
  Sajjadi, Sela, Sitzmann, Stone, Sun, Vora, Wang, Wu, Yi, Zhong, and
  Tagliasacchi]{movi-e:GreffBBDDDFGGHK22}
Klaus Greff, Francois Belletti, Lucas Beyer, Carl Doersch, Yilun Du, Daniel
  Duckworth, David~J. Fleet, Dan Gnanapragasam, Florian Golemo, Charles
  Herrmann, Thomas Kipf, Abhijit Kundu, Dmitry Lagun, Issam~H. Laradji,
  Hsueh{-}Ti~Derek Liu, Henning Meyer, Yishu Miao, Derek Nowrouzezahrai,
  A.~Cengiz {\"{O}}ztireli, Etienne Pot, Noha Radwan, Daniel Rebain, Sara
  Sabour, Mehdi S.~M. Sajjadi, Matan Sela, Vincent Sitzmann, Austin Stone,
  Deqing Sun, Suhani Vora, Ziyu Wang, Tianhao Wu, Kwang~Moo Yi, Fangcheng
  Zhong, and Andrea Tagliasacchi.
\newblock Kubric: {A} scalable dataset generator.
\newblock In \emph{{IEEE/CVF} Conference on Computer Vision and Pattern
  Recognition, {CVPR} 2022, New Orleans, LA, USA, June 18-24, 2022}, pages
  3739--3751. {IEEE}, 2022.
\newblock \doi{10.1109/CVPR52688.2022.00373}.
\newblock URL \url{https://doi.org/10.1109/CVPR52688.2022.00373}.

\bibitem[Gu et~al.(2017)Gu, Lillicrap, Ghahramani, Turner, and
  Levine]{sampleff-metric:gu2017iclr}
Shixiang Gu, Timothy~P. Lillicrap, Zoubin Ghahramani, Richard~E. Turner, and
  Sergey Levine.
\newblock Q-prop: Sample-efficient policy gradient with an off-policy critic.
\newblock In \emph{5th International Conference on Learning Representations,
  {ICLR} 2017, Toulon, France, April 24-26, 2017, Conference Track
  Proceedings}. OpenReview.net, 2017.
\newblock URL \url{https://openreview.net/forum?id=SJ3rcZcxl}.

\bibitem[Hafner et~al.(2020)Hafner, Lillicrap, Ba, and
  Norouzi]{dreamerv1:hafner2020iclr}
Danijar Hafner, Timothy~P. Lillicrap, Jimmy Ba, and Mohammad Norouzi.
\newblock Dream to control: Learning behaviors by latent imagination.
\newblock In \emph{8th International Conference on Learning Representations,
  {ICLR} 2020, Addis Ababa, Ethiopia, April 26-30, 2020}. OpenReview.net, 2020.
\newblock URL \url{https://openreview.net/forum?id=S1lOTC4tDS}.

\bibitem[Hafner et~al.(2021)Hafner, Lillicrap, Norouzi, and
  Ba]{dreamerv2:hafner2021iclr}
Danijar Hafner, Timothy~P. Lillicrap, Mohammad Norouzi, and Jimmy Ba.
\newblock Mastering atari with discrete world models.
\newblock In \emph{9th International Conference on Learning Representations,
  {ICLR} 2021, Virtual Event, Austria, May 3-7, 2021}. OpenReview.net, 2021.
\newblock URL \url{https://openreview.net/forum?id=0oabwyZbOu}.

\bibitem[Hafner et~al.(2023)Hafner, Pasukonis, Ba, and
  Lillicrap]{dreamerv3:hafner2023preprint}
Danijar Hafner, Jurgis Pasukonis, Jimmy Ba, and Timothy~P. Lillicrap.
\newblock Mastering diverse domains through world models.
\newblock \emph{CoRR}, abs/2301.04104, 2023.
\newblock \doi{10.48550/ARXIV.2301.04104}.
\newblock URL \url{https://doi.org/10.48550/arXiv.2301.04104}.

\bibitem[Jia et~al.(2023)Jia, Liu, and Huang]{jia2023iclr}
Baoxiong Jia, Yu~Liu, and Siyuan Huang.
\newblock Improving object-centric learning with query optimization.
\newblock In \emph{The Eleventh International Conference on Learning
  Representations, {ICLR} 2023, Kigali, Rwanda, May 1-5, 2023}. OpenReview.net,
  2023.
\newblock URL \url{https://openreview.net/pdf?id=\_-FN9mJsgg}.

\bibitem[Jiang et~al.(2020)Jiang, Janghorbani, de~Melo, and
  Ahn]{scalor:jiang2020iclr}
Jindong Jiang, Sepehr Janghorbani, Gerard de~Melo, and Sungjin Ahn.
\newblock {SCALOR:} generative world models with scalable object
  representations.
\newblock In \emph{8th International Conference on Learning Representations,
  {ICLR} 2020, Addis Ababa, Ethiopia, April 26-30, 2020}. OpenReview.net, 2020.
\newblock URL \url{https://openreview.net/forum?id=SJxrKgStDH}.

\bibitem[Johnson et~al.(2016)Johnson, Alahi, and
  Fei{-}Fei]{perceptual-loss:johnson2016eccv}
Justin Johnson, Alexandre Alahi, and Li~Fei{-}Fei.
\newblock Perceptual losses for real-time style transfer and super-resolution.
\newblock In Bastian Leibe, Jiri Matas, Nicu Sebe, and Max Welling, editors,
  \emph{Computer Vision - {ECCV} 2016 - 14th European Conference, Amsterdam,
  The Netherlands, October 11-14, 2016, Proceedings, Part {II}}, volume 9906 of
  \emph{Lecture Notes in Computer Science}, pages 694--711. Springer, 2016.
\newblock \doi{10.1007/978-3-319-46475-6\_43}.
\newblock URL \url{https://doi.org/10.1007/978-3-319-46475-6\_43}.

\bibitem[Kabra et~al.(2021)Kabra, Zoran, Erdogan, Matthey, Creswell, Botvinick,
  Lerchner, and Burgess]{simone:kabra2021nips}
Rishabh Kabra, Daniel Zoran, Goker Erdogan, Loic Matthey, Antonia Creswell,
  Matt~M. Botvinick, Alexander Lerchner, and Christopher~P. Burgess.
\newblock Simone: View-invariant, temporally-abstracted object representations
  via unsupervised video decomposition.
\newblock In Marc'Aurelio Ranzato, Alina Beygelzimer, Yann~N. Dauphin, Percy
  Liang, and Jennifer~Wortman Vaughan, editors, \emph{Advances in Neural
  Information Processing Systems 34: Annual Conference on Neural Information
  Processing Systems 2021, NeurIPS 2021, December 6-14, 2021, virtual}, pages
  20146--20159, 2021.
\newblock URL
  \url{https://proceedings.neurips.cc/paper/2021/hash/a860a7886d7c7e2a8d3eaac96f76dc0d-Abstract.html}.

\bibitem[Karpathy(2023)]{nanogpt}
Andrej Karpathy.
\newblock {nanoGPT}: The simplest, fastest repository for training/finetuning
  medium-sized {GPTs} ({Generative} {Pretrained} {Transformers}), 2023.
\newblock URL \url{https://github.com/karpathy/nanoGPT}.

\bibitem[Kingma and Ba(2015)]{adam:kingma2015corr}
Diederik~P. Kingma and Jimmy Ba.
\newblock Adam: {A} method for stochastic optimization.
\newblock In Yoshua Bengio and Yann LeCun, editors, \emph{3rd International
  Conference on Learning Representations, {ICLR} 2015, San Diego, CA, USA, May
  7-9, 2015, Conference Track Proceedings}, 2015.
\newblock URL \url{http://arxiv.org/abs/1412.6980}.

\bibitem[Kipf et~al.(2022)Kipf, Elsayed, Mahendran, Stone, Sabour, Heigold,
  Jonschkowski, Dosovitskiy, and Greff]{savi:kipf22}
Thomas Kipf, Gamaleldin~Fathy Elsayed, Aravindh Mahendran, Austin Stone, Sara
  Sabour, Georg Heigold, Rico Jonschkowski, Alexey Dosovitskiy, and Klaus
  Greff.
\newblock Conditional object-centric learning from video.
\newblock In \emph{The Tenth International Conference on Learning
  Representations, {ICLR} 2022, Virtual Event, April 25-29, 2022}.
  OpenReview.net, 2022.
\newblock URL \url{https://openreview.net/forum?id=aD7uesX1GF\_}.

\bibitem[Kosiorek et~al.(2018)Kosiorek, Kim, Teh, and
  Posner]{sqair:kosiorek2018nips}
Adam~R. Kosiorek, Hyunjik Kim, Yee~Whye Teh, and Ingmar Posner.
\newblock Sequential attend, infer, repeat: Generative modelling of moving
  objects.
\newblock In Samy Bengio, Hanna~M. Wallach, Hugo Larochelle, Kristen Grauman,
  Nicol{\`{o}} Cesa{-}Bianchi, and Roman Garnett, editors, \emph{Advances in
  Neural Information Processing Systems 31: Annual Conference on Neural
  Information Processing Systems 2018, NeurIPS 2018, December 3-8, 2018,
  Montr{\'{e}}al, Canada}, pages 8615--8625, 2018.
\newblock URL
  \url{https://proceedings.neurips.cc/paper/2018/hash/7417744a2bac776fabe5a09b21c707a2-Abstract.html}.

\bibitem[Li et~al.(2022)Li, Zhu, Lei, and Zhang]{physics-causal:li2022aaai}
Zongzhao Li, Xiangyu Zhu, Zhen Lei, and Zhaoxiang Zhang.
\newblock Deconfounding physical dynamics with global causal relation and
  confounder transmission for counterfactual prediction.
\newblock In \emph{Thirty-Sixth {AAAI} Conference on Artificial Intelligence,
  {AAAI} 2022, Thirty-Fourth Conference on Innovative Applications of
  Artificial Intelligence, {IAAI} 2022, The Twelveth Symposium on Educational
  Advances in Artificial Intelligence, {EAAI} 2022 Virtual Event, February 22 -
  March 1, 2022}, pages 1536--1545. {AAAI} Press, 2022.
\newblock \doi{10.1609/AAAI.V36I2.20044}.
\newblock URL \url{https://doi.org/10.1609/aaai.v36i2.20044}.

\bibitem[Lin et~al.(2020)Lin, Wu, Peri, Sun, Singh, Deng, Jiang, and
  Ahn]{space:lin2020iclr}
Zhixuan Lin, Yi{-}Fu Wu, Skand~Vishwanath Peri, Weihao Sun, Gautam Singh, Fei
  Deng, Jindong Jiang, and Sungjin Ahn.
\newblock {SPACE:} unsupervised object-oriented scene representation via
  spatial attention and decomposition.
\newblock In \emph{8th International Conference on Learning Representations,
  {ICLR} 2020, Addis Ababa, Ethiopia, April 26-30, 2020}. OpenReview.net, 2020.
\newblock URL \url{https://openreview.net/forum?id=rkl03ySYDH}.

\bibitem[Locatello et~al.(2020)Locatello, Weissenborn, Unterthiner, Mahendran,
  Heigold, Uszkoreit, Dosovitskiy, and Kipf]{slot-attn:locatello2020nips}
Francesco Locatello, Dirk Weissenborn, Thomas Unterthiner, Aravindh Mahendran,
  Georg Heigold, Jakob Uszkoreit, Alexey Dosovitskiy, and Thomas Kipf.
\newblock Object-centric learning with slot attention.
\newblock In Hugo Larochelle, Marc'Aurelio Ranzato, Raia Hadsell,
  Maria{-}Florina Balcan, and Hsuan{-}Tien Lin, editors, \emph{Advances in
  Neural Information Processing Systems 33: Annual Conference on Neural
  Information Processing Systems 2020, NeurIPS 2020, December 6-12, 2020,
  virtual}, 2020.
\newblock URL
  \url{https://proceedings.neurips.cc/paper/2020/hash/8511df98c02ab60aea1b2356c013bc0f-Abstract.html}.

\bibitem[Loshchilov and Hutter(2019)]{adamw:loshchilov2019iclr}
Ilya Loshchilov and Frank Hutter.
\newblock Decoupled weight decay regularization.
\newblock In \emph{7th International Conference on Learning Representations,
  {ICLR} 2019, New Orleans, LA, USA, May 6-9, 2019}. OpenReview.net, 2019.
\newblock URL \url{https://openreview.net/forum?id=Bkg6RiCqY7}.

\bibitem[McCloskey(1983)]{mccloskey1983intuitive}
Michael McCloskey.
\newblock Intuitive physics.
\newblock \emph{Scientific american}, 248\penalty0 (4):\penalty0 122--131,
  1983.

\bibitem[McCloskey et~al.(1983)McCloskey, Washburn, and
  Felch]{intphys-1:mccloskey1983}
Michael McCloskey, Allyson Washburn, and Linda Felch.
\newblock Intuitive physics: The straight-down belief and its origin.
\newblock \emph{Journal of experimental psychology. Learning, memory, and
  cognition}, 9:\penalty0 636--49, 10 1983.
\newblock \doi{10.1037/0278-7393.9.4.636}.

\bibitem[Micheli et~al.(2023)Micheli, Alonso, and Fleuret]{micheli23}
Vincent Micheli, Eloi Alonso, and Fran{\c{c}}ois Fleuret.
\newblock Transformers are sample-efficient world models.
\newblock In \emph{The Eleventh International Conference on Learning
  Representations, {ICLR} 2023, Kigali, Rwanda, May 1-5, 2023}. OpenReview.net,
  2023.
\newblock URL \url{https://openreview.net/pdf?id=vhFu1Acb0xb}.

\bibitem[Paster et~al.(2021)Paster, McIlraith, and Ba]{paster2021iclr}
Keiran Paster, Sheila~A. McIlraith, and Jimmy Ba.
\newblock Planning from pixels using inverse dynamics models.
\newblock In \emph{9th International Conference on Learning Representations,
  {ICLR} 2021, Virtual Event, Austria, May 3-7, 2021}. OpenReview.net, 2021.
\newblock URL \url{https://openreview.net/forum?id=V6BjBgku7Ro}.

\bibitem[Paszke et~al.(2019)Paszke, Gross, Massa, Lerer, Bradbury, Chanan,
  Killeen, Lin, Gimelshein, Antiga, Desmaison, K{\"{o}}pf, Yang, DeVito,
  Raison, Tejani, Chilamkurthy, Steiner, Fang, Bai, and
  Chintala]{pytorch:paszke2019nips}
Adam Paszke, Sam Gross, Francisco Massa, Adam Lerer, James Bradbury, Gregory
  Chanan, Trevor Killeen, Zeming Lin, Natalia Gimelshein, Luca Antiga, Alban
  Desmaison, Andreas K{\"{o}}pf, Edward~Z. Yang, Zachary DeVito, Martin Raison,
  Alykhan Tejani, Sasank Chilamkurthy, Benoit Steiner, Lu~Fang, Junjie Bai, and
  Soumith Chintala.
\newblock Pytorch: An imperative style, high-performance deep learning library.
\newblock In Hanna~M. Wallach, Hugo Larochelle, Alina Beygelzimer, Florence
  d'Alch{\'{e}}{-}Buc, Emily~B. Fox, and Roman Garnett, editors, \emph{Advances
  in Neural Information Processing Systems 32: Annual Conference on Neural
  Information Processing Systems 2019, NeurIPS 2019, December 8-14, 2019,
  Vancouver, BC, Canada}, pages 8024--8035, 2019.
\newblock URL
  \url{https://proceedings.neurips.cc/paper/2019/hash/bdbca288fee7f92f2bfa9f7012727740-Abstract.html}.

\bibitem[Piloto et~al.(2022)Piloto, Weinstein, Battaglia, and
  Botvinick]{physics-voe:Piloto2022IntuitivePL}
Luis~S. Piloto, Ari Weinstein, Peter~W. Battaglia, and Matthew~M. Botvinick.
\newblock Intuitive physics learning in a deep-learning model inspired by
  developmental psychology.
\newblock \emph{Nature Human Behaviour}, 6:\penalty0 1257 -- 1267, 2022.
\newblock URL \url{https://api.semanticscholar.org/CorpusID:250453635}.

\bibitem[Qi et~al.(2021)Qi, Wang, Pathak, Ma, and
  Malik]{physics-rpin:qi2021iclr}
Haozhi Qi, Xiaolong Wang, Deepak Pathak, Yi~Ma, and Jitendra Malik.
\newblock Learning long-term visual dynamics with region proposal interaction
  networks.
\newblock In \emph{9th International Conference on Learning Representations,
  {ICLR} 2021, Virtual Event, Austria, May 3-7, 2021}. OpenReview.net, 2021.
\newblock URL \url{https://openreview.net/forum?id=\_X\_4Akcd8Re}.

\bibitem[Robine et~al.(2023)Robine, H{\"{o}}ftmann, Uelwer, and
  Harmeling]{robine2023iclr}
Jan Robine, Marc H{\"{o}}ftmann, Tobias Uelwer, and Stefan Harmeling.
\newblock Transformer-based world models are happy with 100k interactions.
\newblock In \emph{The Eleventh International Conference on Learning
  Representations, {ICLR} 2023, Kigali, Rwanda, May 1-5, 2023}. OpenReview.net,
  2023.
\newblock URL \url{https://openreview.net/pdf?id=TdBaDGCpjly}.

\bibitem[Singh et~al.(2022)Singh, Wu, and Ahn]{steve:singh22}
Gautam Singh, Yi{-}Fu Wu, and Sungjin Ahn.
\newblock Simple unsupervised object-centric learning for complex and
  naturalistic videos.
\newblock In Sanmi Koyejo, S.~Mohamed, A.~Agarwal, Danielle Belgrave, K.~Cho,
  and A.~Oh, editors, \emph{Advances in Neural Information Processing Systems
  35: Annual Conference on Neural Information Processing Systems 2022, NeurIPS
  2022, New Orleans, LA, USA, November 28 - December 9, 2022}, 2022.
\newblock URL
  \url{http://papers.nips.cc/paper\_files/paper/2022/hash/735c847a07bf6dd4486ca1ace242a88c-Abstract-Conference.html}.

\bibitem[van~den Oord et~al.(2017)van~den Oord, Vinyals, and
  Kavukcuoglu]{vqvae:oord17}
A{\"{a}}ron van~den Oord, Oriol Vinyals, and Koray Kavukcuoglu.
\newblock Neural discrete representation learning.
\newblock In Isabelle Guyon, Ulrike von Luxburg, Samy Bengio, Hanna~M. Wallach,
  Rob Fergus, S.~V.~N. Vishwanathan, and Roman Garnett, editors, \emph{Advances
  in Neural Information Processing Systems 30: Annual Conference on Neural
  Information Processing Systems 2017, December 4-9, 2017, Long Beach, CA,
  {USA}}, pages 6306--6315, 2017.
\newblock URL
  \url{https://proceedings.neurips.cc/paper/2017/hash/7a98af17e63a0ac09ce2e96d03992fbc-Abstract.html}.

\bibitem[Vaswani et~al.(2017)Vaswani, Shazeer, Parmar, Uszkoreit, Jones, Gomez,
  Kaiser, and Polosukhin]{transformer:vaswani17}
Ashish Vaswani, Noam Shazeer, Niki Parmar, Jakob Uszkoreit, Llion Jones,
  Aidan~N. Gomez, Lukasz Kaiser, and Illia Polosukhin.
\newblock Attention is all you need.
\newblock In Isabelle Guyon, Ulrike von Luxburg, Samy Bengio, Hanna~M. Wallach,
  Rob Fergus, S.~V.~N. Vishwanathan, and Roman Garnett, editors, \emph{Advances
  in Neural Information Processing Systems 30: Annual Conference on Neural
  Information Processing Systems 2017, December 4-9, 2017, Long Beach, CA,
  {USA}}, pages 5998--6008, 2017.
\newblock URL
  \url{https://proceedings.neurips.cc/paper/2017/hash/3f5ee243547dee91fbd053c1c4a845aa-Abstract.html}.

\bibitem[Wu et~al.(2022)Wu, Escontrela, Hafner, Abbeel, and
  Goldberg]{wm-in-robotics:wu2022corl}
Philipp Wu, Alejandro Escontrela, Danijar Hafner, Pieter Abbeel, and Ken
  Goldberg.
\newblock Daydreamer: World models for physical robot learning.
\newblock In Karen Liu, Dana Kulic, and Jeffrey Ichnowski, editors,
  \emph{Conference on Robot Learning, CoRL 2022, 14-18 December 2022, Auckland,
  New Zealand}, volume 205 of \emph{Proceedings of Machine Learning Research},
  pages 2226--2240. {PMLR}, 2022.
\newblock URL \url{https://proceedings.mlr.press/v205/wu23c.html}.

\bibitem[Wu et~al.(2023)Wu, Dvornik, Greff, Kipf, and Garg]{slotformer:wu23}
Ziyi Wu, Nikita Dvornik, Klaus Greff, Thomas Kipf, and Animesh Garg.
\newblock Slotformer: Unsupervised visual dynamics simulation with
  object-centric models.
\newblock In \emph{The Eleventh International Conference on Learning
  Representations, {ICLR} 2023, Kigali, Rwanda, May 1-5, 2023}. OpenReview.net,
  2023.
\newblock URL \url{https://openreview.net/pdf?id=TFbwV6I0VLg}.

\bibitem[Yi et~al.(2020)Yi, Gan, Li, Kohli, Wu, Torralba, and
  Tenenbaum]{clevrer:yi2020iclr}
Kexin Yi, Chuang Gan, Yunzhu Li, Pushmeet Kohli, Jiajun Wu, Antonio Torralba,
  and Joshua~B. Tenenbaum.
\newblock {CLEVRER:} collision events for video representation and reasoning.
\newblock In \emph{8th International Conference on Learning Representations,
  {ICLR} 2020, Addis Ababa, Ethiopia, April 26-30, 2020}. OpenReview.net, 2020.
\newblock URL \url{https://openreview.net/forum?id=HkxYzANYDB}.

\bibitem[Yu et~al.(2023)Yu, Ruan, and Xing]{yu2023ijcai}
Zhongwei Yu, Jingqing Ruan, and Dengpeng Xing.
\newblock Explainable reinforcement learning via a causal world model.
\newblock In \emph{Proceedings of the Thirty-Second International Joint
  Conference on Artificial Intelligence, {IJCAI} 2023, 19th-25th August 2023,
  Macao, SAR, China}, pages 4540--4548. ijcai.org, 2023.
\newblock \doi{10.24963/IJCAI.2023/505}.
\newblock URL \url{https://doi.org/10.24963/ijcai.2023/505}.

\bibitem[Zoran et~al.(2021)Zoran, Kabra, Lerchner, and
  Rezende]{parts:zoran2021iccv}
Daniel Zoran, Rishabh Kabra, Alexander Lerchner, and Danilo~J. Rezende.
\newblock {PARTS:} unsupervised segmentation with slots, attention and
  independence maximization.
\newblock In \emph{2021 {IEEE/CVF} International Conference on Computer Vision,
  {ICCV} 2021, Montreal, QC, Canada, October 10-17, 2021}, pages 10419--10427.
  {IEEE}, 2021.
\newblock \doi{10.1109/ICCV48922.2021.01027}.
\newblock URL
  \url{https://openaccess.thecvf.com/content/ICCV2021/papers/Zoran_PARTS_Unsupervised_Segmentation_With_Slots_Attention_and_Independence_Maximization_ICCV_2021_paper.pdf}.

\end{thebibliography}

}


\clearpage
\appendix

\section{Hyperparameters and configuration}
\label{sec:hypars}

This whole work was implemented in Pytorch \citep{pytorch:paszke2019nips}; the code is open source and published at the following URL: \\
\url{https://github.com/torchipeppo/transformers-and-slot-encoding-for-wm}

\subsection{VQVAE for tokenization}

See \autoref{tab:hypars-vqvae} below. The implementation is based on \citet{taming:esser21} and \citet{micheli23}.

\begin{table}[ht]
\centering
\begin{tabular}{lc}
    \hline
    Hyperparameter & Value \\
    \hline
    Video resolution (pixels) & \(64 \times 64\) \\
    Number of tokens per frame & \(64\) \\
    Channels in convolution & \(64\) \\
    Number of residual conv. layers & \(10\) \\
    Number of self-attention layers & \(3\) \\
    \hline
\end{tabular}
\caption{Hyperparameters for the VQVAE, both encoder and decoder}
\label{tab:hypars-vqvae}
\end{table}

\subsection{Transformers and slot encoding}

See \autoref{tab:hypars-transformers} below. The implementation is based on nanoGPT \citep{nanogpt}. The transformers involved, i.e. the corrector-predictor-decoder triplet and the task success classifier, use the same values for the hyperparameters unless otherwise specified.

\begin{table}[ht]
\centering
\begin{tabular}{lcc}
    \hline
    \multicolumn{2}{l}{Hyperparameter} & Value \\
    \hline
    \multicolumn{2}{l}{Vocabulary size} & \(50304\) \\
    \multicolumn{2}{l}{Number of tokens per frame} & \(64\), as above \\
    \multicolumn{2}{l}{Token embedding dimension} & \(768\) \\
    Number of layers & (corrector) & \(2\) \\
     & (predictor) & \(2\) \\
     & (decoder) & \(6\) \\
     & (task classifier) & \(2\) \\
    \multicolumn{2}{l}{Number of attention heads} & \(12\) \\
    \multicolumn{2}{l}{Number of slots} & \(4\) \\
    \hline
    Given video frames \((N)\) & (task classifier) & \(5\) \\
    \hline
\end{tabular}
\caption{Hyperparameters for the transformer triplet}
\label{tab:hypars-transformers}
\end{table}

\subsection{Training process}

See \autoref{tab:hypars-training}, \ref{tab:hypars-optim-vqvae}, and \ref{tab:hypars-optim-transformers}.

\begin{table}[h!t]
\centering
\begin{tabular}{lcc}
    \hline
    Configuration & & Value \\
    \hline
    \multicolumn{2}{l}{Epochs} & \(100\) \\
    \multicolumn{2}{l}{Batch size \((BS)\)} & \(50\) \\
    \multicolumn{2}{l}{Batches per epoch \((BPE)\)} & \(10\) \\
    \multicolumn{2}{l}{Training steps per epoch} & \(BS \times BPE = 500\) \\
    Data samples & Training & 47500 \\
     & Evaluation & 2500 \\
    \hline
\end{tabular}
\caption{General configuration for the training process}
\label{tab:hypars-training}
\end{table}

\begin{table}[h!t]
\centering
\begin{tabular}{lcc}
    \hline
    Hyperparameter & Value & \\
    \hline
    Type & Adam & \citep{adam:kingma2015corr} \\
    Leaning rate & \(10^{-4}\) & \\
    \hline
\end{tabular}
\caption{Optimizer for the VQVAE}
\label{tab:hypars-optim-vqvae}
\end{table}

\begin{table}[h!t]
\centering
\begin{tabular}{lcc}
    \hline
    Hyperparameter & Value & \\
    \hline
    Type & AdamW & \citep{adamw:loshchilov2019iclr} \\
    Leaning rate & \(6 \cdot 10^{-4}\) & \\
    Weight decay & \(0.1\) & \\
    \( \left( \beta_1, \beta_2 \right) \) & \( \left( 0.9, 0.95 \right) \) & \\
    \hline
\end{tabular}
\caption{Optimizer for each transformer}
\label{tab:hypars-optim-transformers}
\end{table}

\subsection{Classification experiments duration}
\label{sub:classification-duration}

See \autoref{tab:time-classifier}. We remind that the experiments were run on a server with an NVIDIA A100 graphics card, with 40GB memory.

\begin{table}[h!t]
\centering
\begin{tabular}{lcc}
    \hline
    Architecture & Average time (hours) \\
    \hline
    \acronym{} & 1.077 \\
    STEVE & 0.8604 \\
    Decoder-only & 1.999 \\
    \hline
\end{tabular}
\caption{Training time of the classifier for each underlying world modeling architecture.}
\label{tab:time-classifier}
\end{table}

\section{Existing asset attribution}
\label{sec:licenses}

The following implementations have been referenced during this work:

\begin{itemize}
    \item nanoGPT \citep{nanogpt} for the Transformer implementation, licensed under the MIT license;
    \item \citet{taming:esser21} for the VQVAE implementation, released under the MIT license;
    \item \citet{micheli23} for further details about the Transformer and VQVAE implementations, as well as the "decoder only" baseline, licensed under the GPL;
    \item \citet{steve:singh22} for the STEVE baseline, licensed under the MIT license.
\end{itemize}

The PHYRE video dataset has been generated by \citet{physics-rpin:qi2021iclr} from the PHYRE simulator \citep{phyre:bakhtin19}. It was downloaded following the instructions on the author's GitHub repository: \url{https://github.com/HaozhiQi/RPIN/blob/master/docs/PHYRE.md##11-download-our-dataset} \\
Correspondence with the author has confirmed that the dataset is released under the same license as PHYRE itself, i.e. the Apache license.

\end{document}